# Programming of Skill-based Robots

Taneli Lohi
Intelligent Robotics
VTT Technical Research Centre of Finland
Oulu, Finland
taneli.lohi@vtt.fi

Samuli Soutukorva
Intelligent Robotics
VTT Technical Research Centre of Finland
Oulu, Finland
samuli.soutukorva@vtt.fi

Tapio Heikkilä
Intelligent Robotics
VTT Technical Research Centre of Finland
Oulu, Finland
tapio.heikkila@vtt.fi

*Abstract*— **Manufacturing is facing ever changing market demands, with faster innovation cycles resulting to growing agility and flexibility requirements. Industry 4.0 has been transforming the manufacturing world towards digital automation and the importance of software has increased drastically. Easy and fast task programming and execution in robot - sensor systems become a prerequisite for agile and flexible automation and in this paper, we propose such a system. Our solution relies on a robot skill library, which provides the user with high level and parametrized operations, i.e., robot skills, for task programming and execution. Programming actions results to a control recipe in a neutral product context and is based on use of product CAD models or alternatively collaborative use of pointers and tracking sensor with real parts. Practical tests are also reported to show the feasibility of our approach.**

## I. Introduction

Automation in agile and very flexible manufacturing, where lot sizes go at the extreme cases down to one is very demanding, because the set-up and execution of new tasks should be nearly instant. This implies reduced integration effort and easy reuse of available devices and software on the production lines. The devices must offer a uniform interface to fulfil flexibility requirements and new devices and components of a production line must be integrated fast and easy, independent of the component's manufacturer [1]. Software plays a key role, and the transition to small batches implies robot software structures and control architectures that can provide the basis for the required flexibility [2].

Programming robot tasks has traditionally been done by high expertise professionals. Creating robot programs for handling varying and new work objects is time consuming and depending on programming method, may require stopping of production. For small and medium-sized enterprises key challenges of production automation are high costs of expert personnel when small batch sizes require constantly reprogramming of robots [3, 4, 5]. Task level robot programming, relying on parametric robot skills with user friendly and intuitive parameter generation, makes robot programming easier, faster, and more flexible.

To perform successful manipulation, robots depend on precise information about objects in their production environment. In a highly varying environment, such information cannot be given to the robot a priori, and robots must autonomously and continuously acquire information about their surrounding environment to support their decision making and react to unanticipated events [6]. When paired with machine vision systems that detect the work objects and easy robot programming, reusable robot program libraries offer flexibility to manufacturing processes. Accurately calibrated jigs are no more necessary neither are expert personnel for robot programming.

We approach the challenges of agile and flexible robot automation by proposing a system for easy programming and execution of robot tasks. Our solution relies on an executable robot skill library, which provides the user with high level and parametrized operations, i.e., robot skills. Programming is based on use of product CAD models in a neutral format and is carried out using a planning and programming software in the product context, focusing on what operations, i.e., robot skills, need to be carried out to accomplish the required changes in the relations and properties of parts. Additionally, we introduce an interactive option where the programming can be done manually using a physical object and a tracker device. We have developed a planning and programming software system as well as a robot skill library, with which task sequences and all detailed parameters can be easily programmed into a control recipe. The control recipe is interpreted by the control software and skills and operations are allocated to available resources and executed in a robot cell. The skills and synchronization of their possibly parallel executed sequences are specified in detail as UML activity diagrams, the implementation of which as programs is intuitive and straight forward. Our main contribution and novelty is in providing the method and software implementation for easy task planning and programming in the product context – either based on CAD models or interactively using trackers – as well as in the modelling of synchronization of skills. In the following chapter we give a short overview or related work (chapter 2), present our overall architecture (chapter 3), the key principles for programming and control (chapter 4), results from our experimental system (chapter 5) and finally give short conclusions (chapter 6).

## II. Related work

Many robot programming and control systems and architectures have been proposed to provide agility and flexibility for robot automation in small batch production. Task-level programming systems help shop floor workers with low programming knowledge to program robot tasks easier. Tasks can be composed of pre-defined robot skills, and defining parameters for robot skills can be done using Human-Robot Interaction (HRI) [4, 5]. The HRI is typically supported by software with a graphical user interface, but [6] and [7] introduced demonstration-based skill and task programming where skills are parametrized through gestures.

In [4, 8, 9] self-sufficient skills models are introduced for task-level programming. This implies that skills should be parametric, should contain precondition and postcondition checking, and continuous evaluation.

In [10] a task programming scheme is presented where primitive operations are hierarchically combined to create skills, which are sufficient to construct assembly related task-level operations. [11] introduce a skill library consisting of reusable skills which can be incrementally refined by tweaking the parameters of said skills and sequenced to construct assembly related task-level operations. Our skill implementation comply with other skill definitions except not having yet pre- and postconditions.

Task-level approach considers typically three layers of SW components: task, skill, and primitive layers. A task is a high abstraction level operation, that includes a sequence of skills and is used to solve specific goals in robot's workspace [4]. Inspecting quality of a welding seam, or assembling a product are examples for robot tasks. Additionally, skill's target is a physical object in the robot's workspace instead of for example a 3D point. Organizing skill-based systems layers follows the classical three-layer architecture for robot control [13]: at the lowest level there is an interface between the hardware and control module layer and the middle layer, which in turn builds the interface to high-level decision making. The middle layer features sensor-based operation modules such as manipulation, localization, or object recognition, while the third layer consists of a modelling and reasoning components.

A variety of research has been done in CAD, CAM and VRML-based robot planning and programming. However, hardly any of the studies offer intuitive and low-cost robot programming using raw CAD data [12]. A system, allowing users with basic knowledge in CAD and robot programming to generate robot programs from a 3D CAD drawing is presented in [12]. Starting from the CAD model of a robot cell, the user generates a robot program by "drawing" the desired robot paths in the CAD environment. The information needed to program the robot will be extracted from the CAD environment by using an application programming interface (API) provided by the CAD software. This API allows the extraction of the points that characterize each of the different lines used to define a robot path as straight lines, splines, and arcs. However, relying on a vendor specific API restricts the usability of such a system drastically.

One of the most advanced architectural approaches is SkiROS2, a skill-based robot control platform on top of the Robot Operating System (ROS) [2]. SkiROS2 proposes a layered, hybrid control architecture for automated task planning and reactive execution, supported by a knowledge base for reasoning about the world state. The skill formulation based on pre-, hold- and post-conditions, allows to organize robot programs and to compose diverse skills reaching from perception to low-level control and the incorporation of external tools. SkiROS2 proposes automatic programming of skill sequences with a knowledge base and a planner software but lacks the details for setting up and using 3D camera technologies in a flexible manner. In addition, reuse of robot programs implemented in conventional robot languages and executed in robot controllers, is not supported.

ROS is open source and widely used in the academia. In practical applications, large number of ROS nodes are typically run in the same CPU unit. Then ROS may have performance issues. Open-source implementations of OPC UA are also available, but lack the wide application support compared to ROS. Concerning performance, "open62541", has been found as one of the best performing open-source middleware implementations, compared to ROS, and also to MQTT and DDS [14]. ROS2 performs slightly better than ROS1 while using DDS in the transport layer, though with the price of DDS's lower performance under high CPU load and additional message conversions [14, 15].

A hardware-agnostic skill model using the semantic modelling capabilities of OPC UA have been presented in [1]. The model provides a standardized interface to hardware and software functionalities, while offering an intuitive way of grouping multiple skills to a higher hierarchical abstraction. Skill control is following a state machine approach, where a SkillType is represented as a subtype of the OPC UA ProgramStateMachineType. This introduces controllability in a higher fragmentation in time. Still, ways to express the details of synchronization between the skill control modules are not supported. Also, use of robot controllers and conventional robot programs to implement skills are not considered.

In [16] the design and implementation of off-line programmed skills have been introduced, for such as object localization, scanning edges, and straight seam grinding. In [17] agile and ultra-flexible operations lot sizes down to one are considered by utilizing robot skills as services. While the set-up and execution of new tasks must be nearly instant, the focus is on specifying the system architecture and identifying the relevant subsystems and data sources, from which the actual control parameters for robot skills, especially for robot-sensor systems are derived. The feasibility of the approach is shown by experimental tests with intra-logistics and robotic assembly and finalizing tasks. The results reported in this paper are partly relying on the architectural and modelling solutions given in [16] and [17].

III. HIERARCHICAL CONTROL – MODELLING TASKS, SKILLS AND PRIMITIVES

A. Layered architecture

Our control architecture follows the principles of the common three-layer architecture [13]. We have layers for tasks, skills, and primitive operations. Parametrized skills and primitives are specified, designed, implemented, and tested and realized as a reusable component library, relying on which, varying tasks can be planned and programme. Tasks are formulated as sequences of skills, which further on are decomposed according to the behaviour patterns of the skills and primitives. In principle, tasks have no predefined control structures, but are always composed as a sequence of skills, specified by human operator or programmer. Currently, tasks can include only linear sequences of skills, no branching or loops are not yet supported. Skills and primitives have always their predefined sequence control structures, including synchronization of skills and primitives. Planning a task is done by the user by selecting skills one by one from a list of skills and setting a target object for each skill. Further on, the required parameters for skills are derived from the CAD models of the parts connected to the skill, or alternatively on-line by showing from the real objects using a pointer and tracker system. A complete task plan with the skill sequence and related skill parameters is finally composed into a control recipe, which is forwarded to the robot control system. The robot control system merges configuration data of the robot system to finalize the skill and primitive parameters. This approach for the multi layered control system is illustrated in figures 1 and 2.

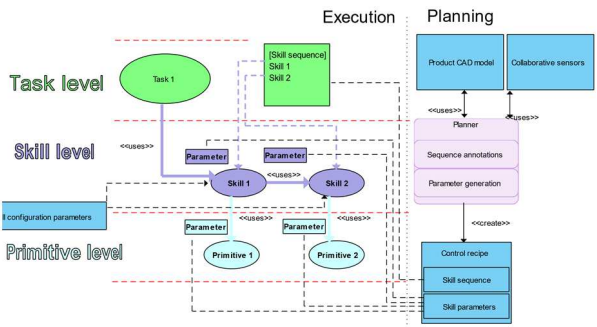

Figure 1. Hierarchical skill-based architecture, in this case a composite skill triggers another skill, and both skills trigger also primitives.

### B. Modelling control sequences

We have the focus on describing synchronized sequence control procedures in three layers (task, skill and primitive operation), in a way which supports easy programming and lays the basis for implementing the multiple control procedures for skill-based robot control. We apply behaviour modelling to model control procedures, in the form of UML /SysML AcitivityActivity Diagrams [18], and in a way illustrating and maintaining the multilayered nature of the control.

Tasks are decomposed into skill sequences. Skill sequences can run parallel, and control flows and synchronization of skill sequences are described in the UML activity models. Each "swim lane" in the activity model represents a lifeline of a control sequence, which can be represented throughout the hierarchy of the three levels. Activity models are seen beneficial because they are intuitively describing the synchronization aspects, and when organized and colour coded, also reflect the control flow in and between the hierarchical layers. In addition, activity models are also easy to be programmed.

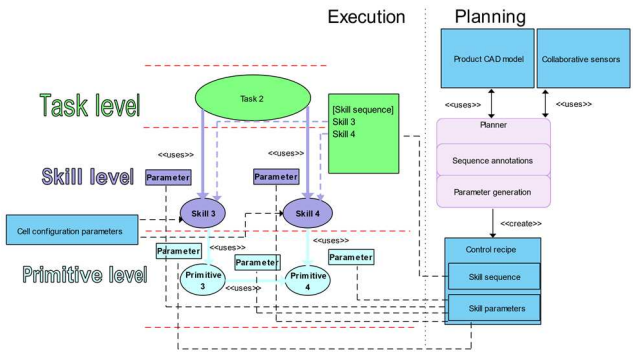

Figure 2. Hierarchical skill-based architecture, in this case both skills trigger primitives, and a composite primitive triggers another primitive.

Figures 1 and 2 illustrate the model of control for a task, with related skills and primitive operations. Tasks are always created as skill sequences and task control triggers the skills. Skill controls trigger primitive operations or other skills, and primitive operations trigger device operations in HW interfaces or other primitive operations. Thus, skills and primitive operations can have a composite structure and trigger entities in the same layer. Skills and primitive operations have starting time input parameters, acquired from the control recipe, and are also represented in the ports in the left side border line of the related activity module. Runtime parameters are represented in the ports of top and bottom-line border lines of the activity. Figure 3 illustrates the control flow from task level to skill and primitive levels, where the models for skill and primitive level control are taken from library specifications - it should be noted that the control flow between skills at the skill level, between skills and primitive operations, and between primitive operations at the primitive level are fixed because they are implemented according to this control model as control programs in the component library. Task descriptions, on the other hand, are created dynamically by the user, although they could be stored in the component library as well.

### C. Implementations of tasks, skills and primitive operations

Each control sequence of a skill or a primitive operation is implemented as a control program component: as an executable, as a module in an executable or as a module in an interpreted control script. Task level control is implemented with an interpreter in the python language, which parses the control recipe and triggers skill controls in the order specified in the control recipe. The control recipe is implemented as a JSON file, which is passed from the planning software to the runtime environment of the task control. Robot cell configuration data is implemented currently embedded in the configuration files of the HW and SW components, or embedded, as global variables in the interpreter programs (see switch/case structure in Figure 4). The task and skill components in Figure 4 are equivalent to the swimlanes in Figure 3 Skill controls are implemented either as python functions calling or sending messages via TCP/IP to other skill controls, or as program blocks in a robot language program code (e.g., KRL, URScript, or RAPID languages). Considering the skill controls and primitive operation controls with a compound structure, library components are implemented as executables, python functions or code blocks in the robot language. This program structure of the control components is illustrated in Figure 4.

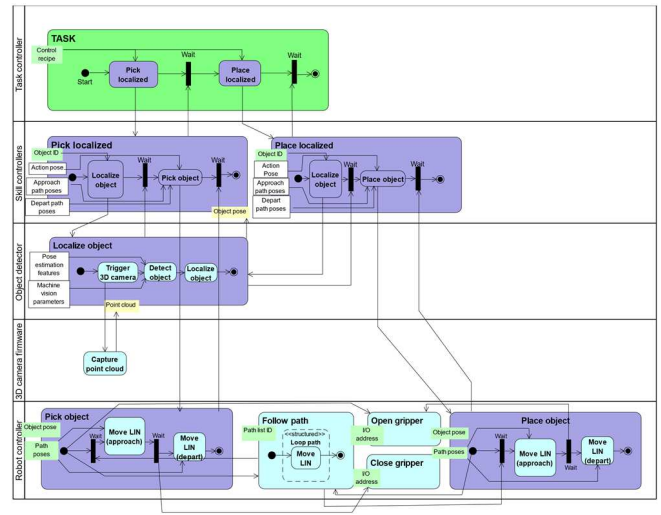

Figure 3. An example of a control flow between layers and within the layers: the task control triggers skill controls, which further trigger other skills or primitive operations (green=task control, purple=skill control, cyan=primitive operation.

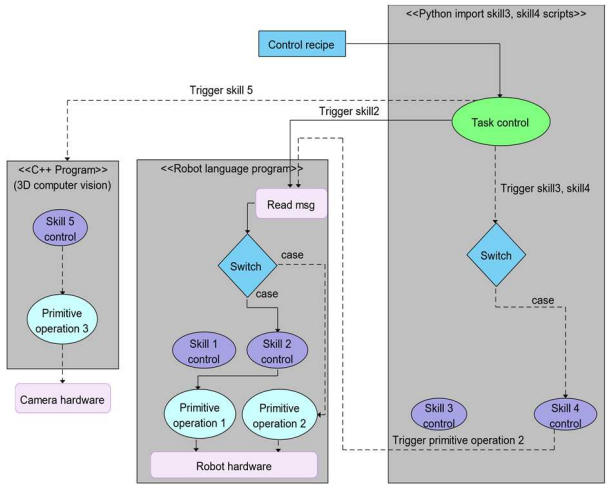

Figure 4. Implementation options for task control, skill control and primitive operation control components: python and robot language scripts and compiled programs.

## IV. PLANNING AND PROGRAMMING SYSTEM FOR TASKS AND SKILLS

### A. CAD based programming

A CAD based programming system has been developed using the geometric engine of the OpenCascade Technology [19]. The operator creates a task by selecting predefined skills from list and adding them to the skill sequence. Each skill is connected to one or more target object to be manipulated during skill execution. For example, in a pick skill: the target object is the object that is picked. A place skill has two target objects: an object to be placed, and another object on top of which the first object is placed. The target objects are selected interactively from visualization of a CAD model and ID of selected part is attached to the skill prototype. Geometric features of the target object to derive parameter values to skills are skill specific.

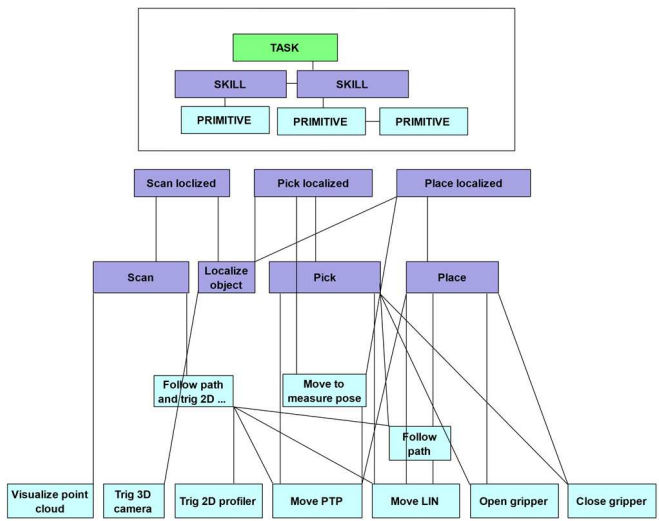

Figure 5. An exemplary set of skills, available for tasks.

Each part in the CAD model has an ID. In an assembly CAD model, the same part can be instantiated multiple times (e.g., a screw). The ID is constructed using the name of the part, name of the part's instance, and name of the assembly: "targetpartname/instancename|assemblyname".

When all skills in the sequence have been given a target, the skills are annotated automatically to the STEP model as a sequence. Each skill annotation contains the order of the skills, a name of the skill and IDs of the target part(s). The annotation is a string, where the ID of the target part is constructed from part, instance, and assembly names. If the skill has multiple target parts, they are separated using a semicolon. While STEP 242 standard supports annotating product information to the model [20], the Open Cascade geometric engine doesn't support writing notes that are attached to a part in a CAD model. Therefore, as a work around, a dimension annotation linked to the part is used as the base for a custom annotation. The annotation text is added as a comment in string format to the dimension annotation.

Currently available skills that are supported by the task planning system, are Pick, Place, Localize object, Scan, Pick localized, Place localized and Scan localized (figure 5). The last three are composite skills that use other skills, e.g., Pick localized skill uses Pick and Localize object skills.

To generate a task control recipe, the skill annotations are read from the STEP file. Each skill is parametrized one-by-one. In the control recipe, e.g., approach-, action- and departure poses for pick and place skills are created using predefined rules. In figure 6 the lower pose is created by selecting four vertices (blue in figure 6). Location of the pose is in the center of gravity of vertices. The orientation is calculated using vectors between vertices. The user interface allows offsetting location and orientation of the generated pose. The upper pose in figure 6 is created by offsetting the lower pose in the pose's y-direction. The task contains at least one skill. A skill has a type, an order number, a part ID and a target ID.

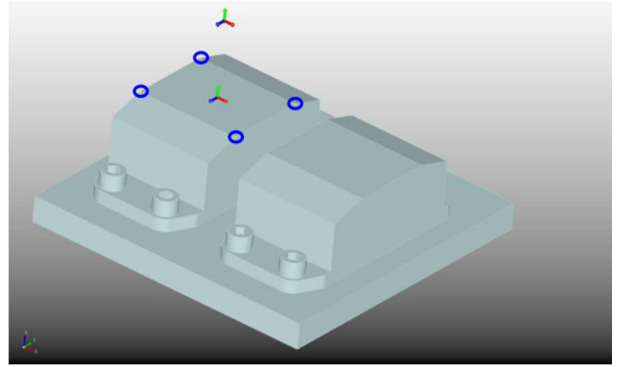

Figure 6. A Grip pose created in middle of plane and other created by offsetting first pose

If the skill is or uses the "Localize object" skill, an additional pose estimation parameter file for our 3D machine vision system is generated. The parameters are calculated using characteristic geometric features of the specific part, that are used to identify and locate the object. The features can be 3D curves like a wire around hole in the object contour, or 3D shaped segments, like spherical or conical. At least three features are needed for full pose estimation. The 3D curves can be a closed chain of edges or a wire. The edges or the wire is sampled with points with fixed distance to each other. Eigen vectors and eigen values are calculated from the point set, using principal component analysis that is provided by the Eigen C++ library [21].

For each feature segment used in pose estimation, a local coordinate system is calculated into the center of gravity of the feature point set of the. A z-axis of the local coordinate system is in the direction of the eigen vector that is linked to the smallest eigen value. The direction of the x-axis is the direction of the eigen vector that is linked to the largest eigen value. Direction of the y-axis is acquired using the cross product between x- and z-axis. Examples of the sampled wires with visualized local coordinate systems are shown in figure 7. The recipe for locating an object contains a type of the feature (e.g., circular 3D curve, conical segment), a principal point, which for 3D curves is the center of gravity of the sampled points, cumulative length of edges or wires in the feature, a local coordinate system, two to four reference points, and list of distances to principal points of other pose estimation features.

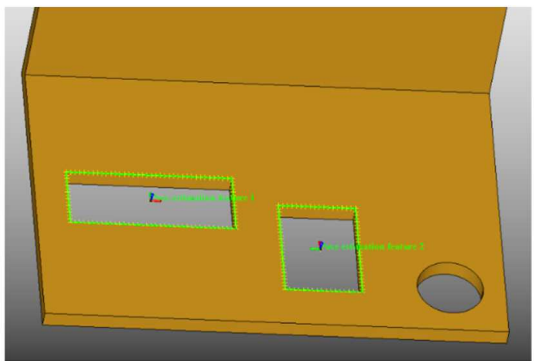

Figure 7. Two pose estimation features with local coordinate systems

### B. Collaborative planning and programming

An alternative method for task programming is to create task descriptions collaboratively in the robot's workspace. A tracked pointer tool is used to measure poses or points from the target work object surfaces, edges or corners. Measured poses are visualized (figure 8), and the poses are added as parameters for a skill.

Collaborative planning uses a 3D-camera with a machine vision software to localize the work object. Coordinate transformations to calculate measured poses or points in the object coordinate system, are shown in figure 9. A 6D tracker with a tracked probe is used to collect 3D-poses or points from a surface of the work object. Coordinate transformations from robot-to-tracker, and robot-to-3D-camera are stored in and acquired from cell configuration files. The transformations are used to convert measured point from the tracker-to-robot and robot-to-camera coordinates system. The user activates the 3D-camera with the machine vision system to localize the work object, and using localization results, the measured poses or points are transformed to the work objects coordinate system. Measured points can be visualized with a 3D model of the target object (figure 8), to show if measurements were successful or not. Finally, the measured or points are set as parameters for a skill.

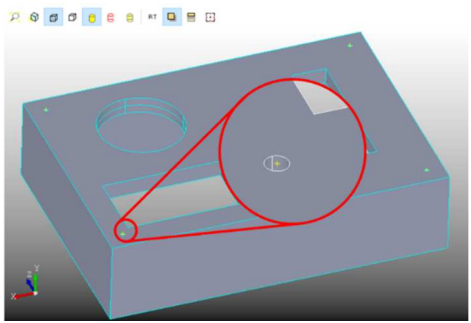

Figure 8. Measured points visualized with CAD model of work object.

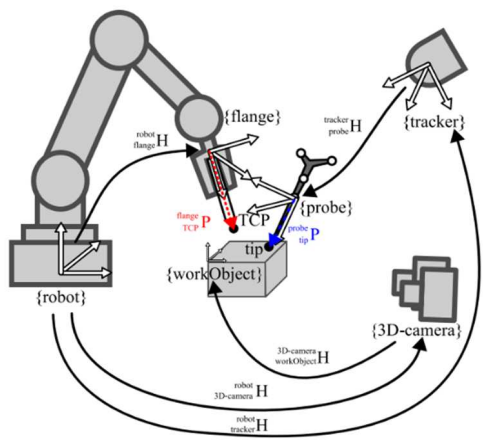

Figure 9. Coordinate frames in collaborative skill programming: teaching the parameter points by showing with a tracking tool.

## V. EXPERIMENTAL SYSTEM AND TESTING

### A. HW/SW architecture

The task programming and control methods were in implemented in our laboratory test facilities. The HW and SW architecture of our test system is illustrated in Figure 10. The software components in Figure 10 utilizes ROS Topic and Services based interfaces to communicate with each other. Each component in Figure 10 is equivalent to swimlane in Figure 3. The CAD based programming SW was implemented as a C++ executable and uses the OpenCascade SW library [19] as a geometric engine. The collaborative programming SW was also implemented as a C++ executable, and is using VTT's 3D computer vision SW and 3D cameras as well as a vision based tracking device with a mechanical pointer [22].

The *skill selector / skill communication manager* was implemented in the used robot's programming language (e.g., KRL, URScript, or RAPID) and it contains the functions for communication and robot controller specific skill sequences and primitives. After a communication initialization phase, where a TCP/IP connection is established with the robot communication interface, the *skill selector / communication manager* enters a switch – case structure where each case corresponds to a sequence control script block, which controls the sequence of actions of a specific skill or a primitive.

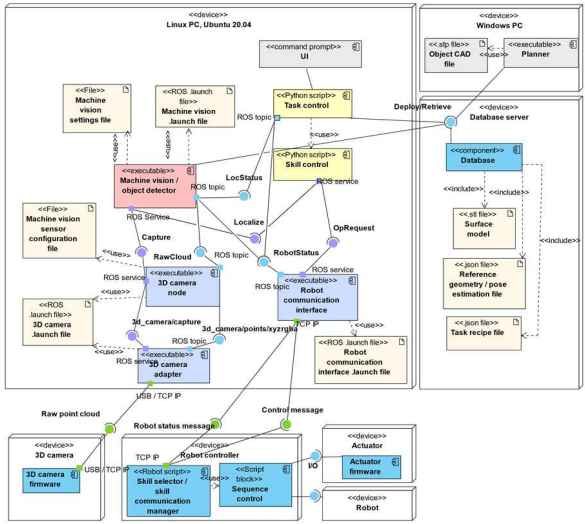

Figure 10. HW/SW implementation architecture.

### B. CAD based task planning and execution

An assembly task was created using a CAD model of the assembly by sequencing skills and attaching related parts to the skills interactively (figure 11, left). The skill sequence was written to STEP model. The skill sequence was read from the model and features that are used to generate parameters for each skill are selected from CAD model of the part (figure 11, right). Control recipes for the task was generated.

The assembly task was executed using the task control, with KUKA Agilus robot equipped with Schmalz suction gripper, and 3D computer vision system with Zivid 2 3D camera. (Figure 12)

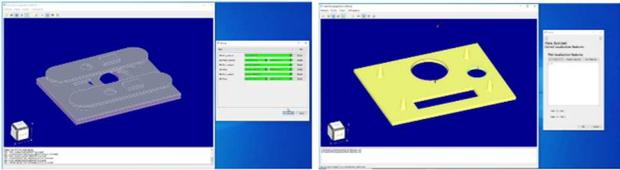

Figure 11. Creating a skill sequence and selecting features for skill parameter generation.

### A. Collaborative task planning and execution

A task was created collaboratively by digitizing points from a work object using 6D tracker. The work object was localized, and the measured points was transformed to the work object's coordinate system. The task recipe with "scan localized" skill was generated.

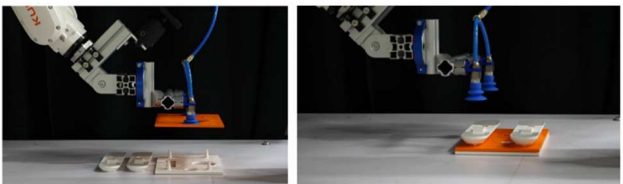

Figure 12. Execution of an assembly task

An experiment system is shown in figure 13. NDI Polaris Vega XT 6D tracker with mechanical pointer was used to digitize points from a work object. Zivid One 3D camera with 3D computer vision system was used to localize the work object. An UR 10e robot with a pointer tool was used to demonstrate scan trajectory.

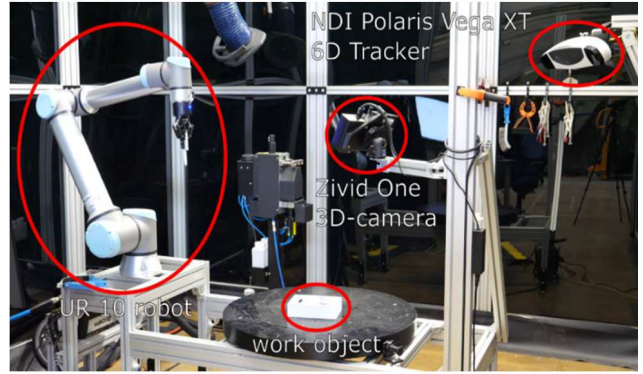

Figure 13. Collaborative task planning and execution experimental system

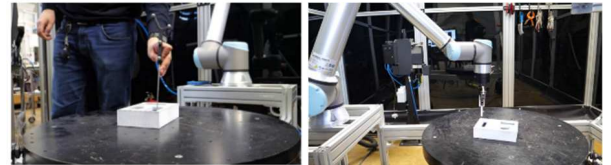

Figure 14. Teaching points for a Scan skill and executing the skill

The tracker was calibrated to the robot's base by attaching a pointer tool to robot's flange, calibrating it as a tool and taking points in the robot and the tracker coordinate system. A transform between was calculated using "least-squares estimation of transformation parameters between two point patterns" [23].

Path points was acquired using the SW with tracked pointer tool (figure 14, left). The work object was localized, and the path points were transformed to the work object's coordinate system. A task recipe, with "scan localized" skill was generated. Parameters for localization skill was generated previously using CAD based task planning software.

Location of the work object was changed, and the task was executed. A pointer tool attached to the robot's flange was used to demonstrate scan trajectory (figure 14, right). At the moment the collaborative task planning is restricted to skills that uses continuous paths.

## VI. CONCLUSIONS

Easy and fast task programming and execution in robot - sensor systems become a prerequisite for agile and flexible automation and in this paper, we propose such a system. Our solution relies on a robot skill library, which provides the user with high level and parametrized operations, i.e., robot skills, for task programming and execution. Programming actions results to a control recipe in a neutral product context and is based on use of product CAD models or alternatively collaborative use of pointers and tracking sensor with real parts. Practical tests are also reported to show the feasibility of our approach.

## VII. DISCUSSION

Our CAD based programming SW is using vendor neutral and open CAD format STEP. Because support for annotation properties of STEP format is supported varyingly and non-uniformly in different CAD products and libraries, the usage of the annotated skills is restricted currently to our software only.

In our skill implementation we use tested primitive operations to compose the skills, which improves skill reliabilities.

Currently the skill control supports only linear skill sequences, where the synchronization of the skills and communication between skills takes place in the task controller and its local variables. In the future, support for other standard control structures like the ones in behaviour trees (e. g., fallback, parallel) will be developed.

Potential applications for task programming and execution system are such as assembly, grinding, polishing, part handling etc. In fact, we see no limitations for applying more widely our approach in industrial robot-sensor systems. Although, experimental validations are showing feasibility of our approach, scalability in large scale production environments still needs to be further studied.

ACKNOWLEDGMENT

The authors would like to thank Mr. Tuomas Seppälä, Mr. Janne Saukkoriipi and Dr. Markku Suomalainen for fruitful discussions and supporting in the use of the skill library. Part of this work was carried out in the INVERSE project, which is funded by the European Union, Horizon Europe research and innovation programme (Grant Agreement 101136067).